\documentclass{article}
\usepackage{ijcai19}

\usepackage{times}
\usepackage{soul}
\usepackage{url}
\usepackage[hidelinks]{hyperref}
\usepackage[utf8]{inputenc}
\usepackage[small]{caption}
\usepackage{graphicx}
\usepackage{amsmath}
\usepackage{booktabs}
\usepackage{natbib}
\usepackage{algorithm2e}
\usepackage{adjustbox}

\let\oldnl\nl
\newcommand{\nonl}{\renewcommand{\nl}{\let\nl\oldnl}} 

\title{Conservative Q-Improvement: Reinforcement Learning for an \newline Interpretable Decision-Tree Policy}

\author{
Aaron M. Roth$^1$\and
Nicholay Topin$^2$\and
Pooyan Jamshidi$^3$\And
Manuela Veloso$^2$\\
\affiliations
$^1$Robotics Institute, Carnegie Mellon University\\
$^2$Machine Learning Department, Carnegie Mellon University\\
$^3$University of South Carolina\\
\emails
\{aaronr1, ntopin\}@andrew.cmu.com,
pjamshid@cse.sc.edu,
mmv@cs.cmu.edu
}

\begin{document}

\maketitle


\begin{abstract}
There is a growing desire in the field of reinforcement learning (and machine learning in general) to move from black-box models toward more ``interpretable AI.'' 
We improve interpretability of reinforcement learning by increasing the utility of decision tree policies learned via reinforcement learning. 
These policies consist of a decision tree over the state space, which requires fewer parameters to express than traditional policy representations.
Existing methods for creating decision tree policies via reinforcement learning focus on accurately representing an action-value function during training, but this leads to much larger trees than would otherwise be required.
To address this shortcoming, we propose a novel algorithm which only increases tree size when the estimated discounted future reward of the overall policy would increase by a sufficient amount.
Through evaluation in a simulated environment, we show that its performance is comparable or superior to traditional tree-based approaches and that it yields a more succinct policy.  Additionally, we discuss tuning parameters to control the tradeoff between optimizing for smaller tree size or for overall reward.
\end{abstract}

\section{Introduction}\label{sec:intro}

Many powerful machine learning algorithms are ``black boxes''~\citep{knight2017dark}.  They produce models which can be used to make predictions or policies that can be used to perform tasks, but these models are often opaque or require significant analysis to penetrate.  ``Black blox'' algorithms are less likely to be accepted for use by an organization, and adoption occurs more slowly than for processes which are directly interpretable~\citep{hihn2015data}. It is difficult for humans to trust decisions that cannot be verified, and thus much research has gone into creating interpretable models~\citep{wu2017interpreting} as part of general growing interest in interpretable AI ~\citep{forbesTrustAI}. 

We look specifically at the field of Reinforcement Learning (RL), which in recent years has demonstrated impressive results on tasks such as the games of Go and Dota2, walking as a simulated human-like agent, and accomplishing a variety of real-world robotic tasks~\citep{gitau_2019,DBLP_survey_RL}.  However, many advanced RL methods today result in policies which perform well but which are not directly interpretable.  We are motivated to build an RL algorithm that can learn a policy that is in a human understandable form. However, in contrast to after-the-fact explanations or approximations, we use an interpretable representation that is simultaneously the policy used to perform the task.

A human-understandable RL model is useful in a scenario where an end-user wishes to continually tweak a policy, or change it after it has been learned. Consider a situation where a household assistant robot learns the preferences of it's owner and how to perform certain kinds of tasks in their home.  If the owner goes on a diet, or otherwise changes some specific aspect of their behavior, it would be useful to communicate this to the robot by simply accessing the robot's settings, selecting the task model, and changing the relevant specific aspect of the policy.  This would be faster than the agent having to learn the simple change over time, and less complicated than requiring the robot to understand a human's request and make the change itself.

To improve the interpretability of RL policies, we focus on learning a decision tree policy.  Decision trees are frequently used to explain processes to humans, such as when a deep network is trained to explain itself~\citep{alaniz2019xoc, huk2018multimodal}.  Decision trees can be represented by text or graphically in a way that a human can readily comprehend.  The tree, once constructed, also has useful information about which states are treated identically and which features yield the highest information gain \citep{buntine1992further, leroux2018information}.
A reasonably-sized tree with labeled attributes enables a human to understand the behavior of a learned policy without having to observe its execution. 

Our main contribution is the Conservative Q-Improvement (CQI) reinforcement learning algorithm. CQI learns a policy in the form of a decision tree, is applicable to any domain in which the state space can be represented as a vector of features, and generates a human-interpretable decision tree policy. Additionally, we evaluate CQI on a modified RobotNav environment \citep{mitchell1993explanation}.

The rest of our paper is organized as follows:  Section~\ref{sec:rw_bg} discusses relevant background.  Section~\ref{sec:rw_rw}, discusses previous related work.  Section~\ref{sec:method} describes our approach.  Section~\ref{sec:env} describes the environment we use to evaluate our method. 
We compare our algorithm to a reference baseline method in Section~\ref{sec:results} and discuss conclusions in Section~\ref{sec:conclusion}.

\section{Background}\label{sec:rw_bg}

\paragraph{Reinforcement Learning}
Reinforcement learning is a field of algorithms for obtaining policies of (state $\rightarrow$ action) mappings through experience in an environment \citep{sutton2018reinforcement}. 
We focus on Q-learning: Given a set of states $s \in S$, set of actions $a \in A$, and an immediate reward for executing an action $a$ in a state $s$, the expected value (Q-value) of performing an action in a state can be learned over time by experimenting with actions in states, observing the reward, and updating the Q-value estimate via the Bellman equation:
\begin{equation}\label{eqn:bellmman}
Q_{t+1}(s,a) \leftarrow (1-\alpha)Q_{t}(s,a) + \alpha ( r + \gamma \space \operatorname*{max}_{a'} Q_{t}(s',a') )
\end{equation}
where $Q_{t}$ is the current Q-value estimate for a state-action pair, $s$ and $a$ are the current state and chosen action, $\alpha$ is the learning rate, $r$ is the immediate reward for choosing action $a$ when in state $s$ (as experienced immediately by interacting with the environment), $s'$ is the next state the agent is in after executing $a$, and $a'$ is the best possible next action, such that the maximum expected future reward from being in state $s'$ is added to $r$ after being discounted by the factor $\gamma$. 

\paragraph{Decision Trees}
Decision trees have often been used to create easy-to-understand solutions to classification problems, among others. They are trees that start at a root node and branch based on conditions~\citep{quinlan1986induction}. 
An additional benefit to decision trees is that they can be represented graphically, which aids in human understanding~\citep{mulrow2002visual}.
Some work has explored combining neural nets and decision tree forms~\citep{tanno2018adaptive, zhao2001evolutionary}, including using decision trees to explain neural nets~\citep{zhang2018interpreting}.
We seek to create a reinforcement learning policy that is in decision tree form.

In an AI context specifically, it has been suggested that huge decision trees are insufficiently comprehensible to humans, and that simplifying trees is important \textit{for humans} to interact with the representation~\citep{quinlan1987simplifying}.  Chess experts had difficulty understanding an early chess-playing algorithm that fully ``explained'' itself with a complex decision tree---too complex a tree becomes opaque~\citep{michie1983inductive, michie1987current}.
It has been shown that, all else being equal, humans prefer simpler explanations compared to more complex ones~\citep{lombrozo2007simplicity}.
Based on the foregoing, we assume for the remainder of the paper that smaller decision trees are more interpretable by humans.

\section{Related Work}\label{sec:rw_rw}
We are not the first to attempt to learn a decision tree policy. The ``G algorithm'' \citep{chapman1991input} learns a decision tree incrementally as new examples are added.  The Lumberjack algorithm ~\citep{uther2000lumberjack} constructs a Linked Decision Forest---like a decision tree but without the need for repeated internal structure which would otherwise occur. 
TTree \citep{uther2003ttree, uther2002tree} solves MDPs or SMDPs and involves both state abstraction (as in our and other methods) and temporal abstraction.  \citep{pyeatt2001decision} is a work which uses a tree as an internal structure and incrementally builds the tree, splitting on Q-values.  Unlike \citep{hester2010generalized}, an approach that uses decision trees as part of the policy representation, we use a single decision tree to represent the entire policy. \citep{wu2012rule} create decision trees using a post-learning transformation, but we instead maintain our policy as a tree at all stages of training and merely update our tree.

The UTree algorithm \citep{mccallum1996reinforcement, uther1998tree} learns a tree, starting with a single abstract state and then continually splitting as appropriate (based on a splitting criterion), ending when a stopping criterion is reached.  This is similar to our method, except that in UTree splits occur based on how accurately the Q-function is represented rather than based on what improves the policy.   
An effort to produce a decision tree via RL is found in \citep{gupta2015policy}, but the final product is not directly interpretable: it consists of a tree with linear Gibbs softmax sub-policies as leaves. In contrast, our approach creates a tree with action choices as leaves.

The best decision-tree-via-RL approach of which we are aware is the Pyeatt Method \citep{pyeatt2003reinforcement} (PM), which we implement and use as a comparison.  PM differs from our method in terms of its splitting criteria and the manner in which splits are performed.
Like our method, PM starts with a single root node and adds branches and nodes over time.  PM maintains a history of changes in Q-values for each leaf, and splits when it believes that this history represents two distributions.  
We instead use a lookahead approach that predicts which split will produce the largest increase in reward.
The trees resulting from the Pyeatt Method tend to be larger than those that result from our approach, which is designed to keep the size of the tree small and manageable.

Indeed, in none of the above cases are there constraints on the size of the decision trees or merit placed on having a smaller tree (beyond gains in performance).
Our work is aimed at enabling 
interpretability; this requires more than simply using a tree structure as an underlying mechanism.
Certainly we want to learn a good policy, but no less important for us is that the resulting form of the policy be easy to understand.

\section{Method}\label{sec:method}

\RestyleAlgo{ruled}
\LinesNumbered
\begin{algorithm}
$Tree \gets$ initial tree is a single leaf node\;
$H_S \gets $ the starting threshold for what potential $\Delta Q$ is required to trigger a split\;
$h_S \gets H_S$\;
$D \gets$ the decay for $H_S$ and $h_S$ values\;
\For{number of episodes}{
    \While{episode not done}{
        $s_t \gets$ current state at timestep $t$\;
        $L \gets$ leaf of $Tree$ corresponding to $s_t$\;
        $a_t, r_t, s_{t+1} \gets$ TakeAction($L$)\;
        UpdateLeafQValues($L, a_t, r_t, s_{t+1}$)\;
        UpdateVisitFrequency($Tree, L$)\;
        UpdatePossibleSplits($L, s_t, a_t, s_{t+1}$)\;
        $bestSplit, bestValue \gets$ BestSplit($Tree, L, a_t$)\;
        \eIf{$bestValue > h_S$}{
            SplitNode($L$, $bestSplit$)\;
        }{
            $h_S \gets h_S \cdot D $
        }
    }
}
 \caption{Conservative Q-Improvement 
 \label{alg:cqi_algo}}
\end{algorithm}

\subsection{Method Summary}\label{sec:method_summary}

We introduce \textbf{Conservative Q-Improvement} (CQI), an algorithm for learning an interpretable decision-tree policy.  CQI learns a policy in the form of a decision tree while only adding to the tree when doing so would improve the policy.

Each node in the tree is a branch node or leaf node. Branch nodes have two children and a condition based on a feature of the state space.  Leaf nodes correspond to abstract states, and indicate actions that are to be taken.

The tree is initialized with a single leaf node, representing a single abstract state.  Our method is strictly additive.  Over time, it creates branches and nodes by replacing existing leaf nodes with a branch node and two child leaf nodes.

The tree is split only when the expected discounted future reward of the new policy would increase above a dynamic threshold. This minimum threshold decreases slowly over the course of training so the split which induces the largest increase in reward is chosen. Additionally, the threshold resets to its initial value after every split. This way, when the current tree does not represent the Q-values as accurately following a change, a greater increase in reward is required to justify a split. In this way, the method is ``conservative'' in performing splits.

The high level algorithm is shown in Algorithm \ref{alg:cqi_algo}.  An overview is given here, with more in-depth explanations following in Section \ref{sec:method_details}.

At a given timestep, the environment will be in state $s_t$.  This will correspond to some leaf node $L$.  When there is a single leaf node, all states correspond to this node.  When there are multiple leaf nodes, the tree must be traversed to determine the corresponding $L$ (each branch node having a boolean condition that operates on $s_t$, indicating which of two children to traverse).

Once $L$ is identified, an action $a_t$ is chosen.  If exploring, a random action is taken.  If exploiting, an action is chosen based on the highest Q-values on that leaf.  Executing the action yields a reward $r_t$ and next-state $s_{t+1}$.  A series of updates are then performed.

The Q-values on leaf $L$ are updated using the standard Bellman equation update.
Each node keeps track of how often it has been visited.
Each leaf node maintains a history of possible splits.  These are potential means of converting the leaf node into a branch node with two child nodes.  These estimated child nodes have hypothetical visit frequencies and Q-values, which are updated at this time.

Then, the possible splits are checked to determine which split would yield the most \textit{split value}, where split value is increase-in-expected-reward moderated by visit frequency.  If this value is beyond a threshold $h_S$, the split occurs.  Otherwise, this threshold is decreased slightly, by a decay value $D$.   In this manner, the split with the best expected reward gain is chosen, and the time between splits is longer when the increase is small, thereby allowing the leaves to better represent the Q-function before a split occurs.  Splits can occur at any timestep.  $h_S$ loosely affects the rate at which splits will occur.

\subsection{Method Details}\label{sec:method_details}

A detailed explanation of the algorithm follows in this section.  The tree itself consists of two node types, ``branching nodes'' and ``leaf nodes''.

A branching node has the following attributes: visit frequency $v$, dimension to split on $m$ (our method assumes a multidimensional state space), and a value to split on $u_b$.  Each state falls within a single node. To determine the node for state $s$, the tree is traversed, starting at the root.  If the value of the state $u_s$ in the dimension $m$ is less than the $u_b$ value on the node, the left child branch of the node is traversed. (If $u_s \geq u_b$, the right branch is traversed.)  When a leaf node is reached, traversal ends---the leaf node $L$ here corresponds to the abstract state $s$.  This is the process referenced in Algorithm \ref{alg:cqi_algo} line 8.

A leaf node has the following attributes: visit frequency $v$, a mapping of actions to Q-values ($Q : (a \rightarrow q)$), and a list of $Split$ objects $Splits$.

Each $Split$ object contains information about potential splits (one per potential split), and has dimension number $m$, value to split on $u_p$, and Q-value for each action and visit frequency for both potential children ($left.q$, $left.v$, $right.q$, and $right.v$).

The TakeAction method in line 9 of Algorithm \ref{alg:cqi_algo} is detailed in Algorithm \ref{alg:takeAction}.  This is the standard RL behavior, where an $\epsilon$ value or function controls whether to explore (randomly choose a valid action) or exploit (use action-Q-value map on $L$ to determine action with highest expected value).  An action, reward, and next state are returned.

\begin{algorithm}
\textbf{TakeAction($L$):}\\
\eIf{$X \sim U$([0,1]) $< \epsilon$}{
    $a_t \gets$ random valid action\;
}{
    $a_t \gets$ the action that has the largest Q-value ($Q(s_t, a_t)$) on $L$\;
}
$r_t \gets$ reward after executing action $a_t$\;
$s_{t+1} \gets$ state after executing action $a_t$\;
Return $a_t, r_t, s_{t+1}$\;
 \caption{Take Action
 \label{alg:takeAction}}
\end{algorithm}

The Q-values on the leaves (for each leaf, one Q-value per valid action) are updated using the standard Bellman update, as shown in Algorithm \ref{alg:updateQValues}.  

\begin{algorithm}
\textbf{UpdateLeafQValues($L$, $a_t$, $r_t$, $s'$):}\\
$L[Q][a_t] \gets (1 - \alpha) L[Q][a_t] + \alpha (r + \gamma \space \operatorname*{max}_{a'} Q(s',a') )$\;
 \caption{UpdateLeafQValues
 \label{alg:updateQValues}}
\end{algorithm}

We keep track of how often a state node is visited. (A node is visited if it is a leaf node corresponding to a state the agent is in, or if it is a parent to such a leaf node.)  In Algorithm \ref{alg:updateVisitFrequency}, these metrics are updated. Note that the update rule corresponds to an exponential average, so it is a natural choice to accompany Q-learning. $N[v]$ indicates the visit frequency of node $N$, and $ \{ a \Longrightarrow b \} : T $ indicates the set of nodes along the direct path from node $a$ to node $b$ in tree $T$. When determining splits later, it is important to weight potential gain by relative frequency visited.  It serves to convert the value increase within a leaf to an estimated policy-wide increase.  The reference to a ``sibling of $N$'' on line 4 refers to the child of a parent of $N$ that is not $N$.  In our method, a node in the tree can have only two children or no children.

\begin{algorithm}
\textbf{UpdateVisitFrequency($Tree$, $L$):}\\
\For{node $N \in \{Tree[root] \Longrightarrow L \} : Tree  $}{
    $N[v] \gets N[v] \cdot d + (1 - d)$\;
    $B \gets$ \text{sibling of} $N$\;
    $B[v] \gets B[v] \cdot d$\;
}
\nonl Hyperparameters: \\
\nonl $d \gets$ visit decay factor
 \caption{UpdateVisitFrequency
 \label{alg:updateVisitFrequency}}
\end{algorithm}

Each leaf keeps track of potential $Splits$ that could be performed (turning a leaf node into branch node and children leaves).  As shown in Algorithm \ref{alg:updatePossibleSplits}, the $Splits$ maintain and update their own metrics for the $potential$ visit frequency and $potential$ Q-values for the case that a specific split is performed. The $\not\sim$ notation is used to express ``opposite side'' Each $split$ has a left and right $side$.  $L[Q]$ refers to the \{action $\rightarrow$ Q-values\} mapping on $L$, and $L[Q][a]$ refers to the Q-value of action $a$ on node $L$.

\begin{algorithm}[b!]
\textbf{UpdatePossibleSplits($L, s_t, a_t, s'$):}\\
\For{$Split \in L[Splits]$}{
    $side \gets $ `left' or `right' depending on which side of the split corresponds to $s_t$\;
    $Split[side][Q][a_t] \gets (1 - \alpha) L[Q][a_t] + \alpha (r + \gamma \space \operatorname*{max}_{a'} Q(s',a') )$\;
    $Split[side][v] \gets Split[side][v] \cdot d + (1 - d)$\;
    $Split[\not\sim side][v] \gets Split[side][v] \cdot d$\;
}
 \caption{UpdatePossibleSplits
 \label{alg:updatePossibleSplits}}
\end{algorithm}

After the updates occur, it is necessary to retrieve information as to which of the potential splits are currently the best possible split, and what the expected gain ($\Delta Q$) would be if that split occurs.  This process is shown in Algorithm \ref{alg:bestSplit}.  In line 2, $V_P$ gets the product of the visit frequencies on all of the nodes along the path from the root node to $L$.  $SQ$ tracks splits and split values, and split values are the sum of the difference between Q-values (on both the left and right side, each weighted by visits).  Each split is investigated to determine which $Split \in L[Splits]$ has the largest $\Delta Q$.  This value along with the split itself are returned to the parent algorithm (where, if $\Delta Q >$ a threshold $h_S$, a split is performed, and if not, $h_S$ is decayed). 

``Performing a split'' means to take a leaf node and turn it into a branch node with two children leaf nodes.  This process is described in Algorithm \ref{alg:splitNode}.

\begin{algorithm}[t!]
\textbf{BestSplit($Tree, L, a_t$):}\\
$V_P \gets \Pi_{N}^{ \{Tree[root] \Longrightarrow L \} : Tree}{(N[v])} $\;
\% $SQ$ is a mapping of splits to split value (indicating expected increase in $Q$ if a split is used)\;
$SQ : (split \rightarrow \Delta Q) \gets \emptyset $\;
\For{$Split \in L[Splits]$}{
    $c_l \gets (\operatorname*{max}_{a'}{(Split[\text{left}][Q][a'])} - L[Q][a_t])$\;
    $c_r \gets (\operatorname*{max}_{a'}{(Split[\text{right}][Q][a'])} - L[Q][a_t])$\;
    $SQ[Split] \gets V_P \cdot ( c_l \cdot Split[\text{left}][v] + c_r \cdot Split[\text{right}][v] )$
}
$bestSplit \gets \operatorname*{arg max}_{split}{SQ[split]}$\;
$bestValue \gets SQ[bestSplit]$\;
 \caption{BestSplit
 \label{alg:bestSplit}}
\end{algorithm}

\begin{algorithm}
\textbf{SplitNode($N$, $bestSplit$):}\\
Let $B$ be an internal branching node.\;
Let $L, R$ be left and right children of $B$, respectively.\;
$B[v] \gets N[v]$\;
$B[m] \gets bestSplit[m]$\;
$B[u] \gets bestSplit[u]$\;
\For{$N \in \{L, R\}$}{
    $N[Splits] \gets$ \{ Add a set of left-right split pairs, one for each dimension and split.\}
}
$L[v] \gets bestSplit[\text{left}][v]$\;
$L[Q] \gets bestSplit[\text{left}][Q]$\;
$R[v] \gets bestSplit[\text{right}][v]$\;
$R[Q] \gets bestSplit[\text{right}][Q]$\;
Return $B$, $L$, $R$\\
\nonl ||||||| \\
\nonl Hyperparameters: \\
\nonl $Q_{init} \gets$ default Q-value\;
\nonl $numSplits \gets$ the number of splits to check for in each dimension\; 
 \caption{SplitNode \newline Splits leaf node $N$ into nodes $B$, $L$, $R$
 \label{alg:splitNode}}
\end{algorithm}


\section{Environment}\label{sec:env}

\begin{figure}[t]
    \centering
    \includegraphics[width=0.45\textwidth]{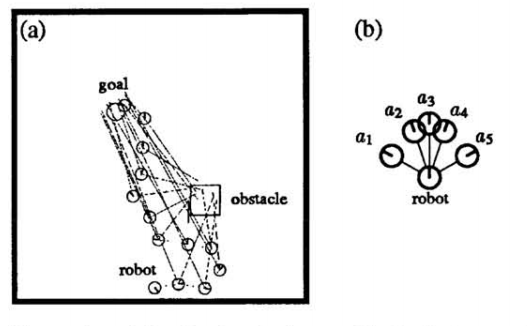}
    \caption{An image of RobotNav, the domain in which we evaluate CQI. As noted in \citep{mitchell1993explanation}: ``a) The simulated robot world, b) Actions''.}
    \label{fig:robot_nav}
\end{figure}

We evaluate our approach on the RobotNav environment from \citep{mitchell1993explanation}. We have also written an open-source OpenAI Gym version of the environment for anyone else to use.\footnote{[Link to code available after review.]}

In this 2D environment, there is a robot, a goal location, and one or more obstacles. The robot must navigate to the goal, avoiding the obstacles, as shown in Figure \ref{fig:robot_nav}.

In addition to the action space in the original work, we also add the option for using the following action set (corresponds to automatically turning to face the goal after every action):
\begin{enumerate}
    \itemsep0cm
    \item move directly towards goal
    \item move directly away from goal
    \item move perpendicular to direct-line-to-goal (right)
    \item move perpendicular to direct-line-to-goal (left)
\end{enumerate}


\section{Results}\label{sec:results}

\subsection{Direct Comparison}

We evaluate the Pyeatt Method and Conservative Q-Improvement Trees on the RobotNav environment. Overall, we find that CQI creates trees with an order of magnitude fewer nodes and greater average reward per episode than the Pyeatt Method. Preliminary experiments on diverse alternative domains exhibit the same behavior.

To fairly compare both methods, we perform a grid search for each method to determine the best hyperparameters. We search over alpha values of 0.005, 0.01, and 0.1 through 0.8 by increments 0.1.  We test hist-mins for the Pyeatt Method of 1000, 3000, 5000, 8000, and 10,000.  For CQI, we try split-thresh-max values from $10^{-2}$ to $10^{9}$ with a step of $\times 10$. We try num-splits values of 2 through 9.  The visit decay and split threshold decay parameters for CQI are held constant at 0.999 and 0.9999, respectively. All of the best hyperparameter configurations fall within the ranges we search.  

The hyperparameters that yield the highest average reward-per-episode for PM are history-list-min-size of 5000 and alpha of 0.3. 
CQI is optimized twice, once for highest reward and once for smallest tree size. We note that optimizing CQI for smallest tree size still yields a policy with higher reward than the Pyeatt Method. The hyperparameters that yield the best reward for CQI are alpha of 0.01, split-thresh-max of 10, and num-splits of 3.  
The hyperparameters that result in the smallest tree size are alpha of 0.2, split-thresh-max of $10^{7}$, and num-splits of 7. 
The results of ten trials for each of these parameter configurations are shown in Table~\ref{tbl:main_results}.  Gamma is set to 0.8 as in \citep{mitchell1993explanation}.  Policies are trained for 500,000 steps and evaluated across 50,000 steps, recording average reward per episode and final tree size. We use epsilon-greedy exploration with parameter $\epsilon = 1$ at the start of training and decay epsilon linearly (to a minimum of $\epsilon = 0.05$) over the course of $100,000$ steps. 

\begin{table}[t!]
  \centering
  \begin{adjustbox}{max width=0.48\textwidth}
\begin{tabular}{l c c c c}
\hline
\textbf{Method}  & \multicolumn{2}{c}{\textbf{Tree Size}} & \multicolumn{2}{c}{\textbf{Avg. Reward Per Episode}} \\
  & Avg. & Std. Dev. & Avg. & Std. Dev \\
\hline
Pyeatt Method & 194.2 & 2.35 & -75.73 & 59.05\\ 
CQI (reward opt) & 20.2 & 1.40 & \textbf{-18.59} & 0.45\\ 
CQI (size opt) & \textbf{7} & 0 & -27.20 & 26.50\\ 
\hline
\end{tabular}
\end{adjustbox}
\caption{Comparison of CQI (optimized for reward or size) to Pyeatt on RobotNav.}\label{tbl:main_results}
\end{table}

CQI outperforms PM while resulting in substantially smaller trees. Additionally, the variance for CQI is smaller.  Even with the best hyperparameters we found, the Pyeatt method's maximum reward at the end of training varied between -182.57 and -19.58. In contrast, we find that CQI (optimized for reward) consistently learns good policies. CQI (optimized for policy size) failed to learn a good policy in one trial, but otherwise performs better than the reward for the best Pyeatt method trial; the average reward across the other nine trials is -18.825 and the variance is 0.47.

We believe that CQI manages to achieve smaller tree size \textit{and} greater reward than the Pyeatt Method because it creates a split only when the policy would improve, not just when the Q-values can be more faithfully represented if a split were added. As a result, CQI avoids extraneous splits, so the agent can benefit from greater abstraction over the state space.

\begin{figure}[b!]
    \centering
    \includegraphics[width=0.45\textwidth]{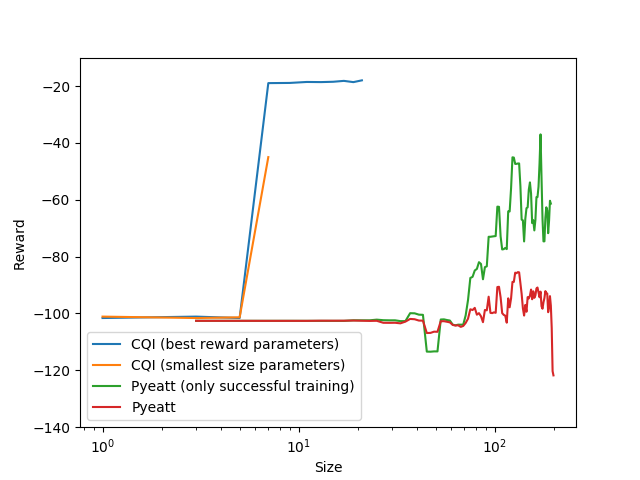}
    \caption{Comparison of Reward vs Tree Size for CQI and Pyeatt on RobotNav.}
    \label{fig:comparison_robotnav}
\end{figure}

We plot the reward obtained by each policy as a function of tree size (calculating reward by evaluating the policy as it existed before the subsequent split) in Figure~\ref{fig:comparison_robotnav}.  The ``only successful runs'' PM line refers to trials where PM achieved more than $-50$ reward by the end of training.  No trials were excluded when plotting the CQI performance.

It is possible to force the Pyeatt method to achieve smaller trees as an end-result of training by increasing history-list-min-size, but it impairs performance.  Increasing this parameter to 50,000 yields trees of size 19 and 21 on RoboNav.  The average reward over 10 trials was -102.62 and -102.59, with variances of 0.80 and 0.01, respectively.  Essentially, when the Pyeatt Method is forced to make trees as small as CQI normally produces, the Pyeatt Method fails to learn.

\subsection{Parameter Sensitivity}

\begin{figure}[t]
    \centering
    \includegraphics[width=0.45\textwidth]{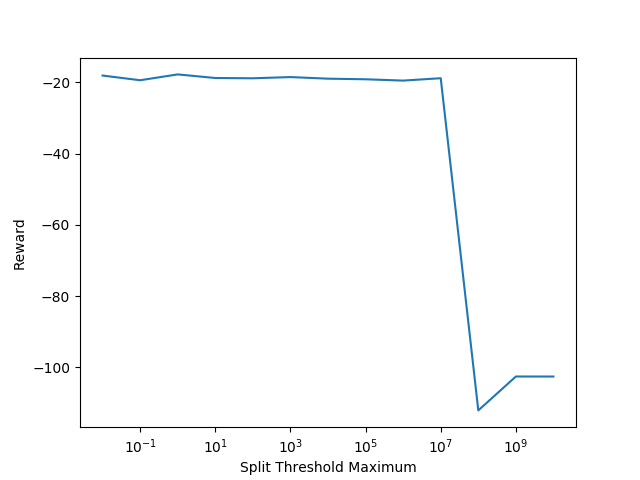}
    \caption{Performance of CQI on RobotNav as a function of split threshold maximum.}
    \label{fig:rn_stm_by_rew}
\end{figure}

We also investigated the sensitivity of CQI to changes in hyperparameter values.  There are a number of hyperparameters that can be adjusted to influence desired behavior.  Optimizing for either size or reward is one such case.  We considered the effect of num-split and split-thresh-max on reward and size, varying each while keeping other parameters constant.  

The num-split parameter has little effect on reward.  Across 10 trials, reward varies only slightly, from -18.84 to -17.88 (excluding one outlier). With other parameters set to the values which minimize tree size, there is no discernible effect on tree size: it ranges between 7 and 9, and with parameters set to values which maximized reward, size ranges from 15 to 21.  (At the same time, during the grid search, we observed a trend of higher num-splits yielding smaller trees for hyperparameter configurations far from those minimizing size or maximizing reward).

\begin{figure}[t]
    \centering
    \includegraphics[width=0.45\textwidth]{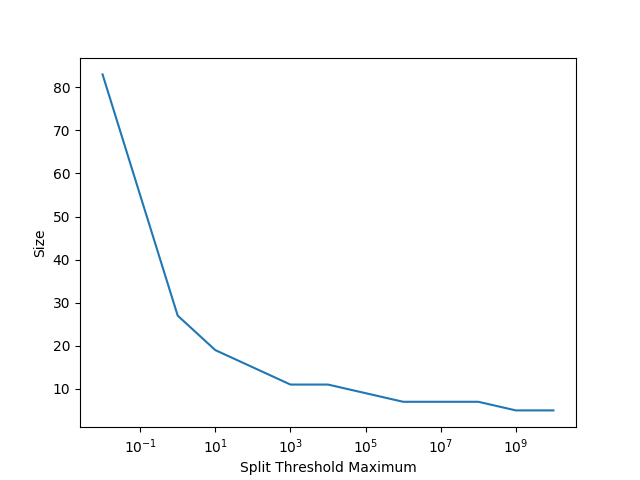}
    \caption{Policy size of CQI on RobotNav as a function of split threshold maximum.}
    \label{fig:rn_stm_by_size}
\end{figure}

Increasing split-threshold-maximum up to $10^7$ has little effect on reward (at parameters otherwise optimized for reward), with final episode reward ranging from -19.49
to -17.75. Note that these are all greater than the best recorded Pyeatt trial which yielded a reward of -19.58. For values greater that $10^8$, the reward drops precipitously, passing below -100, as shown in Figure \ref{fig:rn_stm_by_rew}.  This follows intuition, since split-threshold-maximum controls how long to wait before making a split.  With a fixed training period, too long an interval between splits means that it is impossible to split enough times to create a successful policy.  Another intuitive result of increasing split-threshold-maximum is the very clear trend of decreasing tree size shown in Figure~\ref{fig:rn_stm_by_size}.  Waiting longer to perform a split means that the earlier splits are better informed and therefore higher quality.  When this happens, it is more difficult to improve the policy further, so it takes more steps for potential splits to reach the split threshold.  This result highlights the importance of this parameter since larger values can yield smaller trees.  However, increasing training time allows for smaller trees without sacrificing final reward. As long as the training time is sufficient for the chosen split-threshold-maximum, the final policy will have sufficient time to train and avoid the ``cliff'' which results in poor policies.

\section{Conclusion and Future Work}\label{sec:conclusion}

We have introduced Conservative Q-Improvement.  CQI is a reinforcement learning method applicable to any environment with discrete actions and multidimensional state space.  It produces a policy in the form of a decision tree, and we have designed it to produce smaller trees than existing methods while not sacrificing policy performance.

We investigated the nature of the CQI method, discussing ways that one might optimize for tradeoffs between smaller trees and strictly better policies.  We found that CQI outperformed the Pyeatt method in terms of \textit{both} tree size and reward achieved.  This is due to CQI's conservative nature, whereby it only creates a new node when it will result in an improvement in the policy.

The tradeoffs of CQI can be tuned depending on the use case, and depending on domain.  For example, if there were a domain where even CQI produces large trees, a practitioner could settle for lower reward in order to have a smaller tree that can be understood and inspected.  On the other hand, if all policies for an environment were relatively small, or the policy only had to be manually inspected occasionally or without hurry, then one could choose parameter values which result in a larger, better-performing policy.

Future work could entail tree-building methods that are not strictly additive but rebalance along the way so as to reduce final policy size. Additionally, a regularization term which allows the tree size to directly affect the policy's perceived reward would allow CQI to directly optimize for a trade-off between size and performance on original tasks.



\bibliographystyle{named}
\bibliography{ijcai19}

\begin{thebibliography}{}

\bibitem[\protect\citeauthoryear{Alaniz and Akata}{2019}]{alaniz2019xoc}
Stephan Alaniz and Zeynep Akata.
\newblock Xoc: Explainable observer-classifier for explainable binary
  decisions.
\newblock {\em arXiv preprint arXiv:1902.01780}, 2019.

\bibitem[\protect\citeauthoryear{Arulkumaran \bgroup \em et al.\egroup
  }{2017}]{DBLP_survey_RL}
Kai Arulkumaran, Marc~Peter Deisenroth, Miles Brundage, and Anil~Anthony
  Bharath.
\newblock A brief survey of deep reinforcement learning.
\newblock {\em CoRR}, 2017.

\bibitem[\protect\citeauthoryear{Bloomberg}{2018}]{forbesTrustAI}
Jason Bloomberg.
\newblock Don't trust artificial intelligence? time to open the ai `black box',
  2018.

\bibitem[\protect\citeauthoryear{Buntine and
  Niblett}{1992}]{buntine1992further}
Wray Buntine and Tim Niblett.
\newblock A further comparison of splitting rules for decision-tree induction.
\newblock {\em Machine Learning}, 8(1):75--85, 1992.

\bibitem[\protect\citeauthoryear{Chapman and
  Kaelbling}{1991}]{chapman1991input}
David Chapman and Leslie~Pack Kaelbling.
\newblock Input generalization in delayed reinforcement learning: An algorithm
  and performance comparisons.
\newblock In {\em IJCAI}, volume~91, pages 726--731, 1991.

\bibitem[\protect\citeauthoryear{Gitau}{2019}]{gitau_2019}
Catherine Gitau.
\newblock Success stories of reinforcement learning, 2019.

\bibitem[\protect\citeauthoryear{Gupta \bgroup \em et al.\egroup
  }{2015}]{gupta2015policy}
Ujjwal~Das Gupta, Erik Talvitie, and Michael Bowling.
\newblock Policy tree: Adaptive representation for policy gradient.
\newblock In {\em AAAI}, pages 2547--2553, 2015.

\bibitem[\protect\citeauthoryear{Hester \bgroup \em et al.\egroup
  }{2010}]{hester2010generalized}
Todd Hester, Michael Quinlan, and Peter Stone.
\newblock Generalized model learning for reinforcement learning on a humanoid
  robot.
\newblock In {\em Robotics and Automation (ICRA), 2010 IEEE International
  Conference on}, pages 2369--2374. IEEE, 2010.

\bibitem[\protect\citeauthoryear{Hihn and Menzies}{2015}]{hihn2015data}
Jaitus Hihn and Tim Menzies.
\newblock Data mining methods and cost estimation models: Why is it so hard to
  infuse new ideas?
\newblock In {\em Automated Software Engineering Workshop (ASEW), 2015 30th
  IEEE/ACM International Conference on}, pages 5--9. IEEE, 2015.

\bibitem[\protect\citeauthoryear{Huk~Park \bgroup \em et al.\egroup
  }{2018}]{huk2018multimodal}
Dong Huk~Park, Lisa Anne~Hendricks, Zeynep Akata, Anna Rohrbach, Bernt Schiele,
  Trevor Darrell, and Marcus Rohrbach.
\newblock Multimodal explanations: Justifying decisions and pointing to the
  evidence.
\newblock In {\em Proceedings of the IEEE Conference on Computer Vision and
  Pattern Recognition}, pages 8779--8788, 2018.

\bibitem[\protect\citeauthoryear{Knight}{2017}]{knight2017dark}
Will Knight.
\newblock The dark secret at the heart of ai: no one really knows how the most
  advanced algorithms do what they do-that could be a problem, 2017.

\bibitem[\protect\citeauthoryear{Leroux \bgroup \em et al.\egroup
  }{2018}]{leroux2018information}
Antonin Leroux, Matthieu Boussard, and Remi D{\`e}s.
\newblock Information gain ratio correction: Improving prediction with more
  balanced decision tree splits.
\newblock {\em arXiv preprint arXiv:1801.08310}, 2018.

\bibitem[\protect\citeauthoryear{Lombrozo}{2007}]{lombrozo2007simplicity}
Tania Lombrozo.
\newblock Simplicity and probability in causal explanation.
\newblock {\em Cognitive psychology}, 55(3):232--257, 2007.

\bibitem[\protect\citeauthoryear{McCallum and
  Ballard}{1996}]{mccallum1996reinforcement}
Andrew~Kachites McCallum and Dana Ballard.
\newblock {\em Reinforcement learning with selective perception and hidden
  state}.
\newblock PhD thesis, University of Rochester. Dept. of Computer Science, 1996.

\bibitem[\protect\citeauthoryear{Michie}{1983}]{michie1983inductive}
Donald Michie.
\newblock Inductive rule generation in the context of the fifth generation.
\newblock In {\em Machine Learning Workshop}, page~65, 1983.

\bibitem[\protect\citeauthoryear{Michie}{1987}]{michie1987current}
Donald Michie.
\newblock Current developments in expert systems.
\newblock In {\em Proceedings of the Second Australian Conference on
  Applications of expert systems}, pages 137--156. Addison-Wesley Longman
  Publishing Co., Inc., 1987.

\bibitem[\protect\citeauthoryear{Mitchell and
  Thrun}{1993}]{mitchell1993explanation}
Tom~M Mitchell and Sebastian~B Thrun.
\newblock Explanation-based neural network learning for robot control.
\newblock In {\em Advances in neural information processing systems}, pages
  287--294, 1993.

\bibitem[\protect\citeauthoryear{Mulrow}{2002}]{mulrow2002visual}
Edward~J Mulrow.
\newblock The visual display of quantitative information, 2002.

\bibitem[\protect\citeauthoryear{Pyeatt \bgroup \em et al.\egroup
  }{2001}]{pyeatt2001decision}
Larry~D Pyeatt, Adele~E Howe, et~al.
\newblock Decision tree function approximation in reinforcement learning.
\newblock In {\em Proceedings of the third international symposium on adaptive
  systems: evolutionary computation and probabilistic graphical models},
  volume~2, pages 70--77. Cuba, 2001.

\bibitem[\protect\citeauthoryear{Pyeatt}{2003}]{pyeatt2003reinforcement}
Larry~D Pyeatt.
\newblock Reinforcement learning with decision trees.
\newblock In {\em Applied Informatics}, pages 26--31, 2003.

\bibitem[\protect\citeauthoryear{Quinlan}{1986}]{quinlan1986induction}
J.~Ross Quinlan.
\newblock Induction of decision trees.
\newblock {\em Machine learning}, 1(1):81--106, 1986.

\bibitem[\protect\citeauthoryear{Quinlan}{1987}]{quinlan1987simplifying}
J.~Ross Quinlan.
\newblock Simplifying decision trees.
\newblock {\em International journal of man-machine studies}, 27(3):221--234,
  1987.

\bibitem[\protect\citeauthoryear{Sutton and
  Barto}{2018}]{sutton2018reinforcement}
Richard~S Sutton and Andrew~G Barto.
\newblock {\em Reinforcement learning: An introduction}.
\newblock MIT press, 2018.

\bibitem[\protect\citeauthoryear{Tanno \bgroup \em et al.\egroup
  }{2018}]{tanno2018adaptive}
Ryutaro Tanno, Kai Arulkumaran, Daniel~C Alexander, Antonio Criminisi, and
  Aditya Nori.
\newblock Adaptive neural trees.
\newblock {\em CoRR}, 2018.

\bibitem[\protect\citeauthoryear{Uther and Veloso}{1998}]{uther1998tree}
William~TB Uther and Manuela~M Veloso.
\newblock Tree based discretization for continuous state space reinforcement
  learning.
\newblock In {\em AAAI/IAAI}, pages 769--774, 1998.

\bibitem[\protect\citeauthoryear{Uther and Veloso}{2000}]{uther2000lumberjack}
William~TB Uther and Manuela~M Veloso.
\newblock The lumberjack algorithm for learning linked decision forests.
\newblock In {\em International Symposium on Abstraction, Reformulation, and
  Approximation}, pages 219--232. Springer, 2000.

\bibitem[\protect\citeauthoryear{Uther and Veloso}{2003}]{uther2003ttree}
William~TB Uther and Manuela~M Veloso.
\newblock Ttree: Tree-based state generalization with temporally abstract
  actions.
\newblock In {\em Adaptive agents and multi-agent systems}, pages 260--290.
  Springer, 2003.

\bibitem[\protect\citeauthoryear{Uther}{2002}]{uther2002tree}
William~T Uther.
\newblock Tree-based hierarchical reinforcement learning.
\newblock Technical report, Carnegie Mellon, Univ Pittsburgh, PA, Dept of
  Computer Science, 2002.

\bibitem[\protect\citeauthoryear{Wu \bgroup \em et al.\egroup
  }{2012}]{wu2012rule}
Min Wu, Atsushi Yamashita, and Hajime Asama.
\newblock Rule abstraction and transfer in reinforcement learning by decision
  tree.
\newblock In {\em System Integration (SII), 2012 IEEE/SICE International
  Symposium on}, pages 529--534. IEEE, 2012.

\bibitem[\protect\citeauthoryear{Wu \bgroup \em et al.\egroup
  }{2017}]{wu2017interpreting}
Huijun Wu, Chen Wang, Jie Yin, Kai Lu, and Liming Zhu.
\newblock Interpreting shared deep learning models via explicable boundary
  trees.
\newblock {\em arXiv preprint arXiv:1709.03730}, 2017.

\bibitem[\protect\citeauthoryear{Zhang \bgroup \em et al.\egroup
  }{2018}]{zhang2018interpreting}
Quanshi Zhang, Yu~Yang, Ying~Nian Wu, and Song-Chun Zhu.
\newblock Interpreting cnns via decision trees.
\newblock {\em CoRR}, 2018.

\bibitem[\protect\citeauthoryear{Zhao}{2001}]{zhao2001evolutionary}
Qiangfu Zhao.
\newblock Evolutionary design of neural network tree-integration of decision
  tree, neural network and ga.
\newblock In {\em Evolutionary Computation, 2001. Proceedings of the 2001
  Congress on}, volume~1, pages 240--244. IEEE, 2001.

\end{thebibliography}

\end{document}